# Document Automation Architectures:
# Updated Survey in Light of Large Language Models

Mohammad Ahmadi Achachlouei[1]*, Omkar Patil*, Tarun Joshi, Vijayan N. Nair

Corporate Model Risk, Wells Fargo, USA

This paper surveys the current state of the art in document automation (DA). The objective of DA is to reduce the manual effort during the generation of documents by automatically creating and integrating input from different sources and assembling documents conforming to defined templates. There have been reviews of commercial solutions of DA, particularly in the legal domain, but to date there has been no comprehensive review of the academic research on DA architectures and technologies. The current survey of DA reviews the academic literature and provides a clearer definition and characterization of DA and its features, identifies state-of-the-art DA architectures and technologies in academic research, and provides ideas that can lead to new research opportunities within the DA field in light of recent advances in generative AI and large language models.

CCS CONCEPTS•Applied computing~Document management and text processing~Document preparation•Applied computing~Document management and text processing~Document management•Applied computing~Document management and text processing~Document searching•Applied computing~Document management and text processing~Document capture~Document analysis

## 1 INTRODUCTION

Documents such as legal, technical, and clinical reports are usually highly structured and standardized. They capture evidence and knowledge and convey information necessary for successful business processes that create value for stakeholders [Glushko & McGrath 2008]. However, the manual steps involved in the creation of documents can be very time-consuming, resource-intensive, and prone to human error. Document automation (DA) aims to reduce this manual effort in the document generation process by automatically creating and integrating input from different sources and assembling documents conforming to pre-defined templates. There have been reviews of

---

commercial solutions of DA, particularly in the legal domain [Glaser et al. 2020; Dale 2019; Dale 2020], but to date there has been no comprehensive review of the academic research on DA architectures and technologies. Moreover, recent developments in large language models have caused a paradigm shift in the way text is automatically generated. The implications of these developments on various document automation architectures are considerable. The current survey of DA aims to achieve the following objectives: (i) identify state-of-the-art DA architectures and technologies in academic research; (ii) analyze the current and potential implications of text-generative AI on DA systems; and (iii) specify existing issues and bottlenecks in DA systems that can lead to new research opportunities.

To conduct the current survey, we used Google Scholar to target the peer-reviewed journals and conference proceedings published by Elsevier, IEEE, ACM, Springer, and Taylor & Francis. We also reviewed arXiv preprints related to neural networks. The keywords used in the search comprise combinations of document automation, document assembly, document generation, document engineering, report generation, template construction, natural language generation, and large language models. We identified nearly 500 papers and, after reviewing paper abstracts, selected about 250 papers for further review. At the end, we selected about 100 papers with highest relevance to our survey goals. We classified the DA architectures in the papers and analyzed their implications, applications, and future directions. Section 1 introduces document standards and natural language generation. Section 2 describes the DA architectures and their relation to text-generative AI. Section 3 discusses the main findings and trends of the survey.

**1.1 Definitions and Standards**

The document engineering community defines a document as the union of two components: content and presentation [Gomez et al. 2014]. The document content includes a template that defines the logical structure of the document, plus the components that instantiate the template. The document presentation includes the layout that defines exactly where each piece of content is to be placed and how the piece will appear in the document. Document automation is defined as "the trend of applying software solutions to automate the generation of documents" in the context of legal DA [Glaser et al. 2020].

Digital documents can be represented, stored and exchanged in a variety of formats defined by standardization organizations such as W3C[2] and OASIS [Hackos 2016]. DA systems usually support standard formats such as PDF, TeX (Latex), DOCX, HTML (Hypertext Markup Language), and IPYNB[3] when converting documents from one format to another and generating final documents. DA systems may also utilize standard data-exchange formats such as XML (Extensible Markup Language) and JSON (JavaScript Object Notation) to represent documents in a way which is more machine-readable and more convenient for automated processes.

XML is a standard for defining markup languages with a set of start and end *tags* which can be used to add more information about the main textual content, such as the mode of presentation or semantic information. The structure and allowable elements of an XML document can be defined in an XML DTD (Document Type Definition) file. XML structures can be mapped into HTML, plain text, or other XML structures using an Extensible Style Sheet Transformation (XSLT), which can be used to convert an XML document into other formats recursively. The reviewed studies have also employed XML-based Semantic Web standards [Hitzler 2021], including the following W3C standards:

---

[2] https://www.w3.org/
[3] https://jupyter.org/

1. RDF (Resource Description Framework) is a simple triple-based data model (triple of subject, predicate, object). RDF provides a graph-based formalism for representing metadata.
2. OWL (Web Ontology Language) is a de-facto standard for ontology development. It provides a rich vocabulary to add semantics and context and allow reasoning and inference.
3. SPARQL is an RDF query language able to retrieve and manipulate data stored in RDF.

### 1.2 Classical NLG and Large Language Models

Classical NLG systems, which have offered a reliable and interpretable approach for textual document generation based on structured data, consist of three main modules [Reiter & Dale 2000; Gatt & Krahmer 2018]: (i) content determination (or document planning), which analyzes the signals in the data and determines what messages to convey in the text for a specific audience; (ii) microplanning, which chooses particular words, syntactic constructs, and markup annotations used to communicate the information encoded in the document plan; and (iii) surface realization, which generates the final text through grammatical [Gatt & Reiter 2009], statistical [Bangalore and Rambow 2020] or template [McRoy et al. 2003] methods. An example of generating reports using classical NLG for weather reports in the UK is described in [Sripada et al. 2014].

Language models can generate text based on probabilities of words. Large language models (LLMs) are convenient and their text generation capabilities continue to improve. LLMs can follow instructions better with reinforcement learning [Scheurer et al. 2022] and access external APIs for more context [Schick 2023]. This makes document automation easier and faster with LMs and their tools. LMs can also generate, convert and understand documents from data streams, as shown in section 2.7.

[Deemter et al. 2005] argued that template-based NLG approaches are not necessarily worse than other NLG approaches in terms of maintainability, linguistic strength, and output quality. According to [Dale 2020], most of the commercial data-to-text NLG products in 2020 used templating mechanisms. [Reiter 2016] proposes five levels of sophistication for text generation: from level 1 for template-based approaches to level 5 for dynamically creating documents and controlling the narrative. Recent advances in LLMs have reached the level 5; see the survey by [Zhao 2023] for more details on LLMs.

## 2 DOCUMENT AUTOMATION ARCHITECTURES

### 2.1 Reference Architecture for DA

To present a common vocabulary for DA components and their relationships, we introduce a reference architecture in this section. A key characteristic of DA architectures is the document schema and the related ontology. The schema defines the document structure and tags, while the ontology gives meaning and relationships to the tags. The templates for document automation are based on the document schema. They contain static information and placeholders for variable content from users or external data sources. External data sources can be any data source, such as a database, a language model, a set of rules or semantic data such as RDF triplets. They provide knowledge along with human-generated content for the document. The document fragments are then assembled according to the configuration. The document is processed to handle citations, references, code outputs, etc. Finally, the document is converted to the desired output format. For document analysis, a parser uses the document ontology to parse the input document into RDF triplets or another format. The document can also be parsed using natural language understanding techniques. Storage/version control and authentication can be implemented as needed. Figure 1 shows our proposed reference architecture.

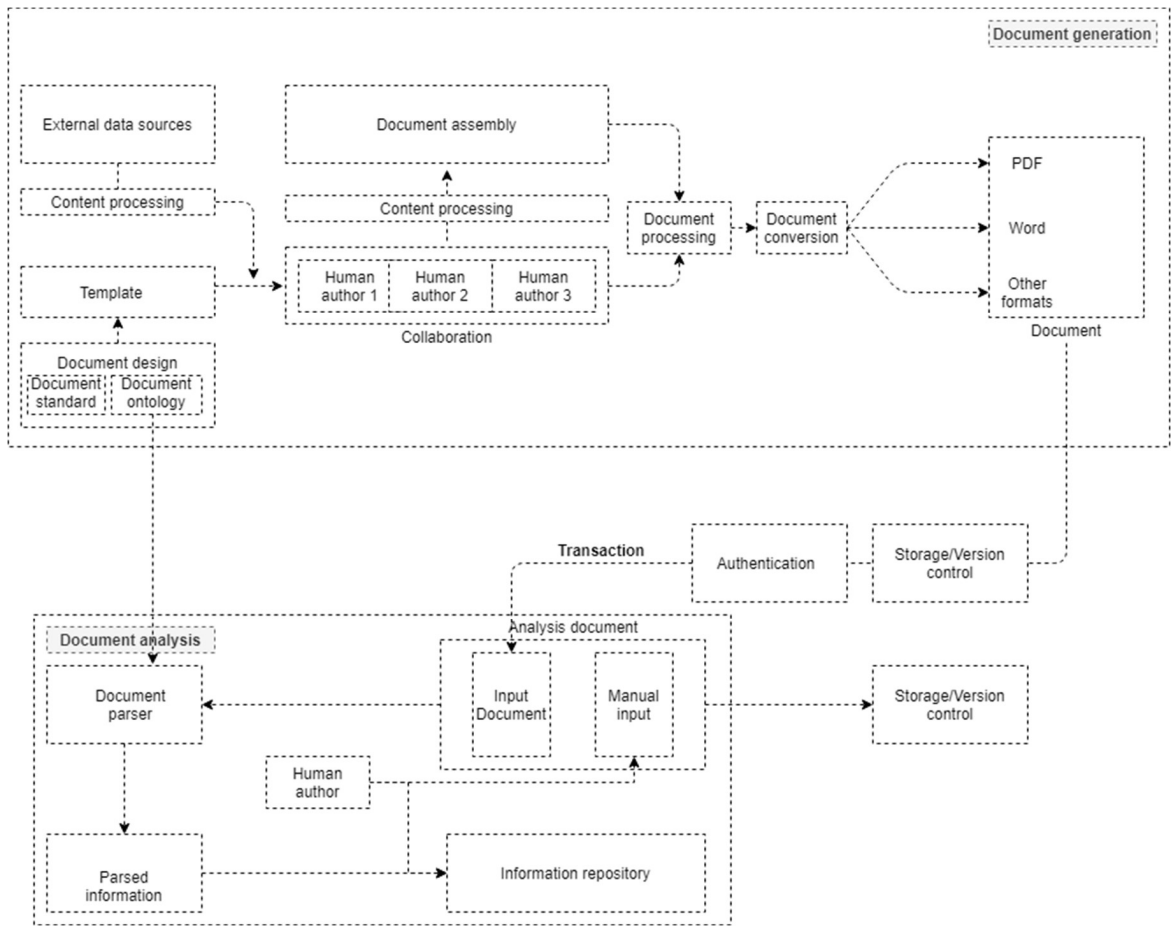

Figure 1. Proposed Reference Architecture for Document Automation

## 2.2 DITA: Darwin Information Typing Architecture

DITA is an XML-based architecture for creating documents based on content reuse, multi-modal output and standardization [Priestley et al. 2001]. DITA suggests that authors write content in small pieces called topics, which can be of three types: tasks, concepts, and references. Topics can be arranged and presented in different ways depending on the purpose and audience of the document. Writing content as topics creates a knowledge graph that lets users navigate the information according to their interest and preference. To make custom markup easier, DITA advises that all custom markup should inherit from a common markup and only define the differences. Creating a document markup means creating a document type definition (DTD) that authors use to write content. The stylesheets can transform the most specific DTD in the hierarchy, so that organizations can create their own output formats by inheriting a general transformation. [Eito-Brun 2020] showed an example of using DITA to improve software documentation in the aerospace industry. DITA enabled efficient integration of data from various sources and single-source generation of HTML/PDF/Word documents. Data from different sources was converted to XML using XSLT. XML files were then assembled and edited using an XML editor. Final documents were produced using an XSLT stylesheet.

### 2.3 DPL: Document Product Line for Variable Content Document Generation

[Gómez et al. 2014] introduced DPLFW, a framework and tool for the Document Product Line (DPL) methodology for multi-user, variable content, and reuse-based document generation. The tool uses a feature model to define the document family and the contributors. Each feature is linked to an info element that can be reused in different documents. The tool generates a document product line that integrates the info elements. Then, the user can select the features they want and get a specific document from the product line. [Penadés et al. 2014] applied DPL to customize tax forms based on user choices.

### 2.4 Model-based Document Generation for Systems Engineering

[Delp et al. 2013] developed a workflow to create customized documents from SysML[4] models in NASA projects. The workflow uses View and Viewpoint concepts to output DocBook files that can be transformed into HTML or PDF. The technique can also be integrated with various analysis tools by using View format as an interface.

[Comoretto et al. 2020] described a document automation solution based on model-based systems engineering (MBSE) in the Rubin observatory. The solution uses MagicDraw[5] to define verification elements and requirements, Jira to report test results, and Docsteady to generate LaTeX test documents from Jira data via REST API. The solution offers three benefits: faster production of verification and validation documents, better integration with the system engineering model, and full traceability of system requirements.

[Michot et al. 2018] proposed an architecture to maintain traceability for bidirectional information transfer between document and the model. Development of APIs along with a tagging mechanism is used to synchronize corresponding elements between the document and its model. There are also other studies on model-to-document generation, such as [Chammard et al. 2020].

### 2.5 Knowledge-Based and Semantic-Web Architectures

[Marković & Gostojić 2020] use an explicit formulation of legal norms prescribing the content and the form of service contracts facilitating the assembly process to generate documents in the Akoma Ntoso format. Akoma Ntoso is an XML vocabulary for legal documents where the mark-up is distributed over the semantic layers of a document to allow for semantic and technical interoperability. [Palmirani & Vitali 2011] define the pillars of the Akoma Ntoso architecture and also present basic elements of Akoma Ntoso XML standard. [Palmirani and Governatori 2018] use Akoma Ntoso for marking up legal text, which then along with legal concepts and rules are used to check GDPR compliance for public sector cloud computing services.

Ontologies such as the Document Components Ontology (DoCO) [Constantin et al. 2016] can enable document automation via enterprise-wide document semantic interoperability. For example, [Mirza & Sah 2017] used an ontology to check for format and structure rules in ACM conference documents. [Colineau et al. 2013] used an ontology to create personalized websites for public administration information delivery. [Pikus et al. 2019] employed an ontology to generate documents from structured data in RDF format. Ontologies can also help integrate data models and information with document automation pipelines. For example, [Colineau et al. 2013] used an ontology verbalizer, called the Semantic Web Authoring Tool [Third et al. 2011], to create query forms that generate personalized websites for public administration users in Australia. [Pikus et al. 2019] utilized RDF data to produce topic-centric and variant-specific documents for lifecycle data.

---

[4] SysML offers two constructs for models: "Viewpoint" refers to the specification which a stakeholder provides for model elements and aspects that they are interested in. "Views" are how the stakeholders see the model according to the viewpoint specified.
[5] https://www.3ds.com/products-services/catia/products/no-magic/magicdraw/

### 2.6 Quantitative Analysis and Clinical Reports

Quantitative analysis reports aim to present reproducible and well-integrated analysis with text and code. Such documentation approaches are usually created on top of the analysis tool or programming languages. For example, Codebraid [Poore 2019] executes code blocks and inline code in Pandoc Markdown documents [MacFarlane 2012] as part of the document build process. Architectures and technologies based on Jupyter notebooks are commonly utilized in scientific and quantitative analysis documentation workflows. Jupyter notebooks provide an interactive web application allowing users to create and share programmatic analysis in over 40 languages and provides features to add data commentary in the same environment [Kluyver et al. 2016; Pérez & Granger 2007].

Jupyter Notebook is a powerful documentation tool with many resources for notebook management and sharing. JupyterHub enables group use of notebooks [Kluyver et al. 2016], Nbviewer[6] allows read-only sharing of notebooks, Binder[7] turns GitHub notebooks into interactive ones, and NbConvert[8] and IPyPublish convert notebooks to other formats (e.g. PDF and HTML) [Sewell 2017].

Arden syntax is a language for clinical event-monitoring using Medical Logic Modules (MLM), which are rules with categories and slots for making clinical decisions. Arden syntax and MLMs can be used to identify and notify clinical events [Hripcsak 1994]. Some studies have extended Arden syntax to generate clinical reports and to create a more general language called PLAIN [Kraus et al. 2016; Kraus 2018; Kraus et al. 2019].

### 2.7 Impact of Large Language Models on DA

Language models have already been used for text generation and document analysis. [Mager et al. 2021] fine-tuned GPT-2 [Radford et al. 2019] to produce text from AMR (abstract meaning representation) by jointly training prediction of the text and reconstruction of AMR. T5 (Text-to-Text Transfer Transformer) by [Raffel et al. 2019] performed impressively on ToTTo [Parikh et al. 2020] for generating text from highlighted cells in a table. However, the generation often involved hallucination and toxicity, making it difficult for general deployment. However, the recent alignment of language models such InstructGPT [Scheurer et al. 2022] and GPT-4 [OpenAI 2023] has been successful in generating text with lower hallucinations and toxicity while being more acceptable to human evaluators than larger language models. Better generation capabilities coupled with ease of use can make these LLMs as the de-facto method of generating text from data or prompts. Among other domains, legal document automation [Lauritsen 2012] has been profoundly impacted by the advent of LLMs. GPT-4 was able to achieve 90th percentile in Uniform Bar Exam [OpenAI 2023] and ChatGPT[9] was able to pass law-school course final exams without human intervention [Choi et al. 2023]. The human-in-the-loop use of language models along with its tooling is expected to gain more adoption in the legal domain [Eloundou et al. 2023].

---

[6] https://github.com/jupyter/nbviewer
[7] https://mybinder.org/
[8] https://nbconvert.readthedocs.io/en/latest/
[9] https://openai.com/blog/chatgpt

Language models have also been used for document analysis, e.g. in [Sifa et al. 2019], which presents a method to automate manual effort in ensuring completeness of financial documents according to legal requirements. With better ability to follow instructions, aligned language models can have significant impact on document analysis; i.e., analysis tasks that used to require engineering can simply be prompted to an LLM. A prominent example of language model usage for document understanding is question answering over a large document using tooling such as LangChain[10]. Moreover, language models can also perform checks or analysis over document text which cannot be easily codified.

Language models also impact knowledge bases and ontologies, which are used for external data sources or document design. OntoGPT by [Caufield et al. 2023] generates ontologies and knowledge bases from text using GPT-4 instruction prompting. On the other hand, studies such as [Chen et al. 2023; Varshney et al. 2023; Li et al. 2022] use ontologies to enhance the quality, or personalize the text generated using language models.

The impact of LLMs on document automation is expected to become more profound as they become better generalists. LLMs, such as OpenAI Codex, have been tuned to be proficient at understanding and generating code. This can not only enable developers to quickly prototype DA technologies, but also potentially allow for processing documents and converting them from one format to another. As LLMs and related tools (e.g., LangChain) improve, it would be possible to draft, process and understand documents within the language model tools.

## 3 DISCUSSION AND CONCLUSION

### 3.1 Key Findings

In this survey, we examine how different domains have adapted the basic document generation workflow to meet their needs and improve their efficiency. Table 1 summarizes the reviewed approaches by their applicable documents, common features, distinct characteristics, and underlying technologies. Moreover, certain emerging or potential research trends were identified as a part of the survey:

1. Personalization is a growing field of research in document automation. It can be done at the content authoring or document assembly stages. For example, users can get personalized tax forms by choosing variability points in the Document Product Line architecture. Likewise, an ontology context model can tailor the text of public communications based on user profile. Language models are promising for document personalization.
2. Some studies use semantic web technologies like RDF and OWL to reason over document content for compliance, auditing and quality assessment. Language models may be more popular for document understanding due to their simplicity.
3. XML/RDF/ontologies are powerful but hard to use and slow to adopt. Jupyter notebooks are popular and fast because they use JSON, which is simple. DA technologies should be user-friendly and have code completion. LLMs can help with prototyping and converting formats. Models like OntoGPT may help generate ontology from text using instruction prompting on GPT-4.
4. To achieve semantic interoperability across the enterprise, an ontology can be created for the document schema if needed. This allows easy data transfer and document editing among departments. Enterprise knowledge graphs can also provide or infer document content.
5. Domain-specific information systems and repositories need better integration with standard documentation technologies. This integration would allow common DA technologies such as XML to easily pull and push data from such repositories, eliminating the manual overhead for users. Language models can help by generating or querying text from knowledge bases. We also need bidirectional traceability between documents and data repositories, so that changes in either side are reflected and tracked in the other (inspired by MDE-based documentation systems).

---

[10] https://langchain.com/

Table 1. Summary and comparison of DA architectures reviewed

| DA approach | Document variety | Common features | Distinct characteristics | Technologies used | Related studies |
| --- | --- | --- | --- | --- | --- |
| Quantitative analysis reports | Statistical analysis and scientific reports | Integrating and presenting analysis code, code output and natural language explanation | Reproducibility, usability | Jupyter Notebooks, Integration of R/S into other markup languages. | 53, 63, 94, 112, 99, 43, 37, 5, 105 |
| Legal document assembly tools | Legal documents, contracts, etc. | Dynamic questionnaire; Capture and reuse inputs; Ability to pull info from databases; Conditional and rule-based logic | Power drafting, template creation and maintainability | Commercial tools: Hotdocs, Contract Express, etc. | 30, 38, 59, 86, 71, 85 |
| Technical documentation | Software/product lifecycle documents | Minimal human intervention, structured reports | Traceability, automatic code documentation | DITA, Programming languages | 20 |
| Model driven engineering | V&V activities: requirements, plan and design, test cases, publish results | Integrating document generation with modeling platform/tool. | Traceability | SysML, XML, DocBook, Latex | 14, 9, 75, 7 |
| Clinical applications of Arden syntax | Documents in daily clinical use such as Discharge letters. | Integration with patient data management systems, templating language should easily integrate with medical domain knowledge | Usability, domain-based, easy to learn. | Arden syntax | 50, 51, 52 |
| Public administration | Documents meant to serve instructions or rules to a large public. | Document assembly is often the focus of such architectures. | Tailored content, variability driven, personalization | Custom software | 73, 8, 88 |
| Data-to-text | Reports based on machine collected and processed data. | No manual authoring. NLG may be involved to convert data to text. | Informative, periodic, natural language generation | NLG | 28, 106, 29, 107 |
| Highly configurable products | Manuals for each configuration of the product. | Assembly of documents according to product configuration is essential. | Reusability, usability, variability identification | DPL, DITA | 95, 32, 88 |
| Semantic web | Usually used where rule bases are common or for personalization or inter-operability. | Usage of ontologies to improve semantic interoperability, knowledge graphs for generating inferences for document assembly. | Semantic web technology is used. | XML, Custom software | 71, 117, 77, 8, 92, 119 |

Note: The "related studies" column lists the reviewed studies related to each DA approach. To view the reviewed studies, please refer to Appendix A.

## 3.2 Conclusion

We surveyed different document automation architectures for various domains and their goals, such as reusability, interoperability, reproducibility, traceability, personalization and variability of content. We analyzed how each architecture element can benefit from the recent advances in LLMs. We foresee that language models and supporting tools will play a bigger role in document creation and understanding. We also suggest that combining language models with document assembly methods can enhance generative and compositional capabilities.

**REFERENCES**

Achachlouei, M.A., Patil, O., Joshi, T. and Nair, V.N., 2021. Document Automation Architectures and Technologies: A Survey. *arXiv preprint* arXiv:2109.11603.

Bangalore, S. and Rambow, O., 2000. Exploiting a probabilistic hierarchical model for generation. In COLING 2000 Volume 1: The 18th International Conference on Computational Linguistics.

Chammard, T.B., Regalia, B., Karban, R. and Gomes, I., 2020. Assisted authoring of model-based systems engineering documents. In Proceedings of the 23rd ACM/IEEE International Conference on Model Driven Engineering Languages and Systems: Companion Proceedings (pp. 1-7).

Chen, Z., Liu, Y., Chen, L., Zhu, S., Wu, M. and Yu, K., 2022. OPAL: Ontology-Aware Pretrained Language Model for End-to-End Task-Oriented Dialogue. arXiv preprint arXiv:2209.04595.

Choi, J.H., Hickman, K.E., Monahan, A. and Schwarcz, D., 2023. ChatGPT goes to law school. Available at SSRN.

Colineau, N., Paris, C. and Vander Linden, K., 2013. Automatically producing tailored web materials for public administration. New Review of Hypermedia and Multimedia, 19(2), pp.158-181.

Comoretto, G., Guy, L., O'Mullane, W., Bechtol, K., Carlin, J.L., Van Klaveren, B., Roberts, A. and Sick, J., 2020. Documentation automation for the verification and validation of Rubin Observatory software. In Modeling, Systems Engineering, and Project Management for Astronomy IX (Vol. 11450, p. 114500E). International Society for Optics and Photonics.

Constantin, A., Peroni, S., Pettifer, S., Shotton, D. and Vitali, F., 2016. The document components ontology (DoCO). Semantic Web, 7(2), pp.167-181.

Dale, R., 2019. Law and word order: NLP in legal tech. Natural Language Engineering, 25(1), pp.211-217.

Dale, R., 2020. Natural language generation: The commercial state of the art in 2020. Natural Language Engineering, 26(4), pp.481-487.

Deemter, K.V., Theune, M. and Krahmer, E., 2005. Real versus template-based natural language generation: A false opposition?. Computational Linguistics, 31(1), pp.15-24.

Delp, C., Lam, D., Fosse, E. and Lee, C.Y., 2013. Model based document and report generation for systems engineering. In 2013 IEEE Aerospace Conference (pp. 1-11). IEEE.

Eito-Brun, R., 2020. An Automated Pipeline for the Generation of Quality Reports. In European Conference on Software Process Improvement (pp. 706-714). Springer, Cham.

Eloundou, T., Manning, S., Mishkin, P. and Rock, D., 2023. Gpts are gpts: An early look at the labor market impact potential of large language models. arXiv preprint arXiv:2303.10130.

Gatt, A. and Reiter, E., 2009. SimpleNLG: A realisation engine for practical applications. In Proceedings of the 12th European Workshop on Natural Language Generation (ENLG 2009) (pp. 90-93).

Gatt, A. and Krahmer, E., 2018. Survey of the state of the art in natural language generation: Core tasks, applications and evaluation. Journal of Artificial Intelligence Research, 61, pp.65-170.

Glaser, I., Huynh, T., Klymenko, O., Labrenz, B. and Matthes, F., 2020. Legal Document Automation Tool Survey 2020. The Technical University of Munich. Munich, Germany.

Glushko, R.J. and McGrath, T., 2008. Document Engineering: Analyzing and Designing Documents for Business Informatics and Web Services. MIT Press. Cambridge, MA.

Gómez, A., Penadés, M.C., Canós, J.H., Borges, M.R. and Llavador, M., 2014. A framework for variable content document generation with multiple actors. Information and Software Technology, 56(9), pp.1101-1121.

Hackos, J.T., 2016. International standards for information development and content management. IEEE Transactions on Professional Communication, 59(1), pp.24-36.

Hitzler, P., 2021. A review of the semantic web field. Communications of the ACM, 64(2), pp.76-83.

Hripcsak, G., 1994. Writing Arden Syntax medical logic modules. Computers in Biology and Medicine, 24(5), pp.331-363.

Kluyver, T., Ragan-Kelley, B., Pérez, F., Granger, B.E., Bussonnier, M., Frederic, J., Kelley, K., Hamrick, J.B., Grout, J., Corlay, S. and Ivanov, P., 2016. Jupyter Notebooks-a publishing format for reproducible computational workflows. In Proceedings of the 20th International Conference on Electronic Publishing (Vol. 2016, pp. 87-90).

Kraus, S., Castellanos, I., Albermann, M., Schuettler, C., Prokosch, H.U., Staudigel, M. and Toddenroth, D., 2016. Using Arden Syntax for the Generation of Intelligent Intensive Care Discharge Letters. In MIE (pp. 471-475).

Kraus, S., 2018. Generalizing the Arden Syntax to a Common Clinical Application Language. In MIE (pp. 675-679).

Kraus, S., Toddenroth, D., Unberath, P., Prokosch, H.U. and Hueske-Kraus, D., 2019. An Extension of the Arden Syntax to Facilitate Clinical Document Generation. Studies in Health Technology and Informatics, 259, pp.65-70.

Lauritsen, M., 2012. Document Automation. Online course on "Topics in Digital Law Practice" by Center for Computer-Assisted Legal Instruction.

Li, F., UK, A., Hogg, D.C. and Cohn, A.G., 2022, November. Ontology Knowledge-enhanced In-Context Learning for Action-Effect Prediction. In Advances in Cognitive Systems. ACS-2022.

MacFarlane, J., 2012. Pandoc User's Guide. www.pandoc.org (Accessed 15.5.2021).

Mager, M., Astudillo, R.F., Naseem, T., Sultan, M.A., Lee, Y.S., Florian, R. and Roukos, S., 2020. GPT-too: A language-model-first approach for AMR-to-text generation. arXiv preprint arXiv:2005.09123.

Marković, M. and Gostojić, S., 2020. A knowledge-based document assembly method to support semantic interoperability of enterprise information systems. Enterprise Information Systems, pp.1-20.

McRoy, S.W., Channarukul, S. and Ali, S.S., 2003. An augmented template-based approach to text realization. Natural Language Engineering, 9(4), p.381.

Michot, A., Ponsard, C. and Boucher, Q., 2018. Towards Better Document to Model Synchronisation: Experimentations with a Proposed Architecture. In MODELSWARD (pp. 567-574).

Mirza, A.R. and Sah, M., 2017. Automated software system for checking the structure and format of ACM SIG documents. New Review of Hypermedia and Multimedia, 23(2), pp.112-140.

OpenAI, 2023. GPT-4 Technical Report. arXiv preprint arXiv:2303.08774

Palmirani, M. and Governatori, G., 2018, December. Modelling Legal Knowledge for GDPR Compliance Checking. IN JURIX (pp. 101-110).

Palmirani, M. and Vitali, F., 2011. Akoma-Ntoso for legal documents. In Legislative XML for the semantic Web (pp. 75-100). Springer, Dordrecht.

Parikh, A.P., Wang, X., Gehrmann, S., Faruqui, M., Dhingra, B., Yang, D. and Das, D., 2020. Totto: A controlled table-to-text generation dataset. arXiv preprint arXiv:2004.14373.

Penadés, M.C., Martí, P., Canós, J.H. and Gómez, A., 2014. Product Line-based customization of e-Government documents. In PEGOV 2014: Personalization in e-Government Services, Data and Applications (Vol. 1181). CEUR-WS.

Pérez, F. and Granger, B.E., 2017. The state of Jupyter: How Project Jupyter got here and where we are headed. Available at: https://www.oreilly.com/radar/the-state-of-jupyter (Accessed 15.5.2021).

Pikus, Y., Weißenberg, N., Holtkamp, B. and Otto, B., 2019. Semi-automatic ontology-driven development documentation: generating documents from RDF data and DITA templates. In Proceedings of the 34th ACM/SIGAPP Symposium on Applied Computing (pp. 2293-2302).

Poore, G.M., 2019. Codebraid: Live Code in Pandoc Markdown.

Priestley, M., Hargis, G. and Carpenter, S., 2001. DITA: An XML-based technical documentation authoring and publishing architecture. Technical communication, 48(3), pp.352-367.

Radford, A., Wu, J., Child, R., Luan, D., Amodei, D. and Sutskever, I., 2019. Language models are unsupervised multitask learners. OpenAI blog, 1(8), p.9.

Raffel, C., Shazeer, N., Roberts, A., Lee, K., Narang, S., Matena, M., Zhou, Y., Li, W. and Liu, P.J., 2019. Exploring the limits of transfer learning with a unified text-to-text transformer. arXiv preprint arXiv:1910.10683.

Reiter, E. and Dale, R., 2000. Building Natural Language Generation Systems. Cambridge University Press.

Reiter. E., 2016. NLG vs Templates: Levels of Sophistication in Generating Text. https://ehudreiter.com/2016/12/18/nlg-vs-templates/ (Accessed 15.5.2021).

Scheurer, J., Campos, J.A., Chan, J.S., Chen, A., Cho, K. and Perez, E., 2022. Training language models with natural language feedback. arXiv preprint arXiv:2204.14146.

Schick, T., Dwivedi-Yu, J., Dessì, R., Raileanu, R., Lomeli, M., Zettlemoyer, L., Cancedda, N. and Scialom, T., 2023. Toolformer: Language models can teach themselves to use tools. arXiv preprint arXiv:2302.04761.

Sewell, C., 2017, IPyPublish: A package for creating and editing publication ready scientific reports and presentations from Jupyter Notebooks. https://ipypublish.readthedocs.io/en/latest/ (Accessed 15.5.2021).

Sifa, R., Ladi, A., Pielka, M., Ramamurthy, R., Hillebrand, L., Kirsch, B., Biesner, D., Stenzel, R., Bell, T., Lübbering, M. and Nütten, U., 2019. Towards automated auditing with machine learning. In Proceedings of the ACM Symposium on Document Engineering 2019 (pp. 1-4).

Sripada, S., Burnett, N., Turner, R., Mastin, J. and Evans, D., 2014, June. A case study: NLG meeting weather industry demand for quality and quantity of textual weather forecasts. In Proceedings of the 8th International Natural Language Generation Conference (INLG) (pp. 1-5).

Third, A., Williams, S. and Power, R., 2011. OWL to English: a tool for generating organised easily-navigated hypertexts from ontologies. In 10th International Semantic Web Conference (ISWC 2011), 23-27 Oct 2011, Bonn, Germany.

Varshney, D., Zafar, A., Behera, N.K. and Ekbal, A., 2023. Knowledge grounded medical dialogue generation using augmented graphs. Scientific Reports, 13(1), p.3310.

Zhao, W.X., Zhou, K., Li, J., Tang, T., Wang, X., Hou, Y., Min, Y., Zhang, B., Zhang, J., Dong, Z. and Du, Y., 2023. A Survey of Large Language Models. arXiv preprint arXiv:2303.18223.

# APPENDIX A: REVIEWED STUDIES LISTED IN TABLE 1

[1] Bangalore, S. and Rambow, O., 2000. Exploiting a probabilistic hierarchical model for generation. In COLING 2000 Volume 1: The 18th International Conference on Computational Linguistics.

[2] Barbosa, D., Mendelzon, A., Keenleyside, J. and Lyons, K., 2002. ToXgene: a template-based data generator for XML. In Proceedings of the 2002 ACM SIGMOD International Conference on Management of Data.

[3] Bavaresco, R., Silveira, D., Reis, E., Barbosa, J., Righi, R., Costa, C., Antunes, R., Gomes, M., Gatti, C., Vanzin, M. and Silva, E., 2020. Conversational agents in business: A systematic literature review and future research directions. Computer Science Review, 36, p.100239.

[4] Bateman, J.A., 1997. Enabling technology for multilingual natural language generation: the KPML development environment. Natural Language Engineering, 3(1), pp.15-55.

[5] Beg, M., Taka, J., Kluyver, T., Konovalov, A., Ragan-Kelley, M., Thiéry, N.M. and Fangohr, H., 2021. Using Jupyter for reproducible scientific workflows. Computing in Science & Engineering, 23(2), pp.36-46.

[6] Brown, T.B., Mann, B., Ryder, N., Subbiah, M., Kaplan, J., Dhariwal, P., Neelakantan, A., Shyam, P., Sastry, G., Askell, A. and Agarwal, S., 2020. Language models are few-shot learners. arXiv preprint arXiv:2005.14165.

[7] Chammard, T.B., Regalia, B., Karban, R. and Gomes, I., 2020. Assisted authoring of model-based systems engineering documents. In Proceedings of the 23rd ACM/IEEE International Conference on Model Driven Engineering Languages and Systems: Companion Proceedings (pp. 1-7).

[8] Colineau, N., Paris, C. and Vander Linden, K., 2013. Automatically producing tailored web materials for public administration. New Review of Hypermedia and Multimedia, 19(2), pp.158-181.

[9] Comoretto, G., Guy, L., O'Mullane, W., Bechtol, K., Carlin, J.L., Van Klaveren, B., Roberts, A. and Sick, J., 2020. Documentation automation for the verification and validation of Rubin Observatory software. In Modeling, Systems Engineering, and Project Management for Astronomy IX (Vol. 11450, p. 114500E). International Society for Optics and Photonics.

[10] Cote, M., Rezvanifar, A. and Albu, A.B., 2020. Automatic Generation of Electrical Plan Documents from Architectural Data. In Proceedings of the ACM Symposium on Document Engineering 2020 (pp. 1-4).

[11] Dale, R., 2019. Law and word order: NLP in legal tech. Natural Language Engineering, 25(1), pp.211-217.

[12] Dale, R., 2020. Natural language generation: The commercial state of the art in 2020. Natural Language Engineering, 26(4), pp.481-487.

[13] Deemter, K.V., Theune, M. and Krahmer, E., 2005. Real versus template-based natural language generation: A false opposition?. Computational Linguistics, 31(1), pp.15-24.

[14] Delp, C., Lam, D., Fosse, E. and Lee, C.Y., 2013. Model based document and report generation for systems engineering. In 2013 IEEE Aerospace Conference (pp. 1-11). IEEE.

[15] Devlin, J., Chang, M.W., Lee, K. and Toutanova, K., 2018. Bert: Pre-training of deep bidirectional transformers for language understanding. arXiv preprint arXiv:1810.04805.

[16] Dominici, M., 2014. An overview of Pandoc. TUGboat, 35(1), pp.44-50.

[17] Dušek, O. and Jurcicek, F., 2015. Training a natural language generator from unaligned data. In Proceedings of the 53rd Annual Meeting of the Association for Computational Linguistics and the 7th International Joint Conference on Natural Language Processing (Volume 1: Long Papers) (pp. 451-461).

[18] Dušek, O. and Jurčíček, F., 2016. Sequence-to-sequence generation for spoken dialogue via deep syntax trees and strings. arXiv preprint arXiv:1606.05491.

[19] Dušek, O., Novikova, J. and Rieser, V., 2020. Evaluating the state-of-the-art of end-to-end natural language generation: The e2e nlg challenge. Computer Speech & Language, 59, pp.123-156.

[20] Eito-Brun, R., 2020. An Automated Pipeline for the Generation of Quality Reports. In European Conference on Software Process Improvement (pp. 706-714). Springer, Cham.

[21] Elhadad, M. and Robin, J., 1996. An overview of SURGE: A reusable comprehensive syntactic realization component.

[22] El-Kassas, W.S., Salama, C.R., Rafea, A.A. and Mohamed, H.K., 2020. Automatic text summarization: A comprehensive survey. Expert Systems with Applications, p.113679.

[23] Feng, Z., 2013. Functional Grammar and Its Implications for English Teaching and Learning. English Language Teaching, 6(10), pp.86-94.

[24] Ferreira, T.C., Calixto, I., Wubben, S. and Krahmer, E., 2017. Linguistic realisation as machine translation: Comparing different MT models for AMR-to-text generation. In Proceedings of the 10th International Conference on Natural Language Generation (pp. 1-10).

[25] Fielding, R.T., 2000. Architectural styles and the design of network-based software architectures (Vol. 7). Irvine: University of California, Irvine.

[26] Gardent, C., Shimorina, A., Narayan, S. and Perez-Beltrachini, L., 2017. The WebNLG challenge: Generating text from RDF data. In Proceedings of the 10th International Conference on Natural Language Generation (pp. 124-133).

[27] Gatt, A. and Reiter, E., 2009. SimpleNLG: A realisation engine for practical applications. In Proceedings of the 12th European Workshop on Natural Language Generation (ENLG 2009) (pp. 90-93).

[28] Gatt, A. and Krahmer, E., 2018. Survey of the state of the art in natural language generation: Core tasks, applications and evaluation. Journal of Artificial Intelligence Research, 61, pp.65-170.

[29] Gatt, A., Portet, F., Reiter, E., Hunter, J., Mahamood, S., Moncur, W. and Sripada, S., 2009. From data to text in the neonatal intensive care unit: Using NLG technology for decision support and information management. AI Communications, 22(3), pp.153-186.

[30] Glaser, I., Huynh, T., Klymenko, O., Labrenz, B. and Matthes, F., 2020. Legal Document Automation Tool Survey 2020. The Technical University of Munich. Munich, Germany.

[31] Glushko, R.J. and McGrath, T., 2008. Document Engineering: Analyzing and Designing Documents for Business Informatics and Web Services. MIT Press. Cambridge, MA.

[32] Gómez, A., Penadés, M.C., Canós, J.H., Borges, M.R. and Llavador, M., 2014. A framework for variable content document generation with multiple actors. Information and Software Technology, 56(9), pp.1101-1121.

[33] Gong, H., Bi, W., Feng, X., Qin, B., Liu, X. and Liu, T., 2020. Enhancing Content Planning for Table-to-Text Generation with Data Understanding and Verification. In Proceedings of the 2020 Conference on Empirical Methods in Natural Language Processing: Findings (pp. 2905-2914).

[34] Gottesman, B., 2008. Introduction to Meaning-Text Theory, http://www.coli.uni-saarland.de/~tania/CMGD/Ben.Gottesman.pdf (Accessed 15.5.2021).

[35] Hackos, J.T., 2016. International standards for information development and content management. IEEE Transactions on Professional Communication, 59(1), pp.24-36.

[36] Halliday, M.A.K. and Matthiessen, C.M., 2013. Halliday's introduction to functional grammar. Routledge.

[37] Henderson, M.L., Krinsman, W., Cholia, S., Thomas, R. and Slaton, T., 2019. Accelerating Experimental Science Using Jupyter and NERSC HPC. In Tools and Techniques for High Performance Computing (pp. 145-163). Springer.

[38] Henley, B.K., 2020. Document Assembly: What It Is and How to Evaluate Competing Programs, Law Practice, 46, p.30.

[39] Hitzler, P., 2021. A review of the semantic web field. Communications of the ACM, 64(2), pp.76-83.

[40] Höhenberger, S. and Scholta, H., 2017. Will Government Forms Ever be Consistent? Detecting Violations in Form Structures by Utilizing Graph Theory.

[41] Hripcsak, G., 1994. Writing Arden Syntax medical logic modules. Computers in Biology and Medicine, 24(5), pp.331-363.

[42] Ittoo, A. and van den Bosch, A., 2016. Text analytics in industry: Challenges, desiderata and trends. Computers in Industry, 78, pp.96-107.

[43] Juneau, S., Olsen, K., Nikutta, R., Jacques, A. and Bailey, S., 2021. Jupyter-enabled astrophysical analysis using data-proximate computing platforms. Computing in Science & Engineering, 23(2), pp.15-25.

[44] Kahane, S., 1984. The meaning-text theory.

[45] Kale, M., 2020. Text-to-text pre-training for data-to-text tasks. arXiv preprint arXiv:2005.10433.

[46] Kasper, R.T., 1989. A flexible interface for linking applications to PENMAN's sentence generator. In Speech and Natural Language: Proceedings of a Workshop Held at Philadelphia, Pennsylvania, February 21-23, 1989.

[47] Kay, M., 1984. Functional unification grammar: A formalism for machine translation. In 10th International Conference on Computational Linguistics and 22nd Annual Meeting of the Association for Computational Linguistics (pp. 75-78).

[48] Kim, S., Haug, P.J., Rocha, R.A. and Choi, I., 2008. Modeling the Arden Syntax for medical decisions in XML. International Journal of Medical Informatics, 77(10), pp.650-656.

[49] Kluyver, T., Ragan-Kelley, B., Pérez, F., Granger, B.E., Bussonnier, M., Frederic, J., Kelley, K., Hamrick, J.B., Grout, J., Corlay, S. and Ivanov, P., 2016. Jupyter Notebooks-a publishing format for reproducible computational workflows. In Proceedings of the 20th International Conference on Electronic Publishing (Vol. 2016, pp. 87-90).

[50] Kraus, S., Castellanos, I., Albermann, M., Schuettler, C., Prokosch, H.U., Staudigel, M. and Toddenroth, D., 2016. Using Arden Syntax for the Generation of Intelligent Intensive Care Discharge Letters. In MIE (pp. 471-475).

[51] Kraus, S., 2018. Generalizing the Arden Syntax to a Common Clinical Application Language. In MIE (pp. 675-679).

[52] Kraus, S., Toddenroth, D., Unberath, P., Prokosch, H.U. and Hueske-Kraus, D., 2019. An Extension of the Arden Syntax to Facilitate Clinical Document Generation. Studies in Health Technology and Informatics, 259, pp.65-70.

[53] Lang, D.T., 2001. Embedding S in other languages and environments. In Proceedings of DSC (Vol. 2, p. 1).

[54] Langkilde, I. and Knight, K., 1998. The practical value of n-grams is in generation. In Natural Language Generation.

[55] Langkilde, I., 2000. Forest-based statistical sentence generation. In 1st Meeting of the North American Chapter of the Association for Computational Linguistics.

[56] Lankester, R., 2018. Implementing Document Automation: Benefits and Considerations for the Knowledge Professional. Legal Information Management, 18(2), pp.93-97.

[57] Lau, S. and Hug, J., 2018. Nbinteract: Generate interactive web pages from Jupyter notebooks. Technical Report No. UCB/EECS-2018-57, Electrical Engineering and Computer Sciences, University of California at Berkeley.

[58] Lauritsen, M., 2007. Current frontiers in legal drafting systems. In Proceedings of the 11th International Conference on AI and Law.

[59] Lauritsen, M., 2012. Document Automation. Online course on "Topics in Digital Law Practice" by Center for Computer-Assisted Legal Instruction.

[60] Lauritsen, M. and Steenhuis, Q., 2019. Substantive Legal Software Quality: A Gathering Storm?. In Proceedings of the Seventeenth International Conference on Artificial Intelligence and Law (pp. 52-62).

[61] Lavoie, B. and Rainbow, O., 1997. A fast and portable realizer for text generation systems. In Fifth Conference on Applied Natural Language Processing (pp. 265-268).

[62] Lehtonen, M., Petit, R., Heinonen, O. and Lindén, G., 2002. A dynamic user interface for document assembly. In Proceedings of the 2002 ACM

Symposium on Document Engineering (pp. 134-141).

[63] Leisch, F., 2002. Sweave: Dynamic generation of statistical reports using literate data analysis. In Compstat (pp. 575-580). Physica, Heidelberg.

[64] Lewis, M., Liu, Y., Goyal, N., Ghazvininejad, M., Mohamed, A., Levy, O., Stoyanov, V. and Zettlemoyer, L., 2019. Bart: Denoising sequence-to-sequence pre-training for natural language generation, translation, and comprehension. arXiv preprint arXiv:1910.13461.

[65] Li, J., Tang, T., Zhao, W.X. and Wen, J.R., 2021a. Pretrained Language Models for Text Generation: A Survey. arXiv preprint arXiv:2105.10311.

[66] Li, J., Tang, T., Zhao, W.X., Wei, Z., Yuan, N.J. and Wen, J.R., 2021b. Few-shot knowledge graph-to-text generation with pretrained language models. arXiv preprint arXiv:2106.01623.

[67] Lipton, Z.C., Vikram, S. and McAuley, J., 2015. Generative concatenative nets jointly learn to write and classify reviews. arXiv preprint arXiv:1511.03683.

[68] MacFarlane, J., 2012. Pandoc User's Guide. www.pandoc.org (Accessed 15.5.2021).

[69] Mager, M., Astudillo, R.F., Naseem, T., Sultan, M.A., Lee, Y.S., Florian, R. and Roukos, S., 2020. GPT-too: A language-model-first approach for AMR-to-text generation. arXiv preprint arXiv:2005.09123.

[70] Mann, W.C., 1983, August. An overview of the Penman text generation system. In AAAI (pp. 261-265).

[71] Marković, M. and Gostojić, S., 2020. A knowledge-based document assembly method to support semantic interoperability of enterprise information systems. Enterprise Information Systems, pp.1-20.

[72] McRoy, S.W., Channarukul, S. and Ali, S.S., 2003. An augmented template-based approach to text realization. Natural Language Engineering, 9(4), p.381.

[73] Menezes, J.A., da Silva, A.R. and de Sousa Saraiva, J., 2019. Citizen-Centric and Multi-Curator Document Automation Platform: the Curator Perspective. In the 28th International Conference on Information Systems Development (ISD2019).

[74] Mi, L., Li, C., Du, P., Zhu, J., Yuan, X. and Li, Z., 2018. Construction and application of an automatic document generation model. In 2018 26th International Conference on Geoinformatics (pp. 1-6). IEEE.

[75] Michot, A., Ponsard, C. and Boucher, Q., 2018. Towards Better Document to Model Synchronisation: Experimentations with a Proposed Architecture. In MODELSWARD (pp. 567-574).

[76] Minnen, G., Carroll, J. and Pearce, D., 2001. Applied morphological processing of English. Natural Language Engineering, 7(3), pp.207-223.

[77] Mirza, A.R. and Sah, M., 2017. Automated software system for checking the structure and format of ACM SIG documents. New Review of Hypermedia and Multimedia, 23(2), pp.112-140.

[78] Monshi, M.M.A., Poon, J. and Chung, V., 2020. Deep learning in generating radiology reports: A survey. Artificial Intelligence in Medicine, p.101878.

[79] Motahari, H., Duffy, N., Bennett, P. and Bedrax-Weiss, T., 2021. A Report on the First Workshop on Document Intelligence (DI) at NeurIPS 2019. ACM SIGKDD Explorations Newsletter, 22(2), pp.8-11.

[80] Nicolás, J. and Toval, A., 2009. On the generation of requirements specifications from software engineering models: A systematic literature review. Information and Software Technology, 51(9), pp.1291-1307.

[81] Nielsen, J., 1994. Enhancing the explanatory power of usability heuristics. In Proceedings of the SIGCHI conference on Human Factors in Computing Systems (pp. 152-158).

[82] Nurseitov, N., Paulson, M., Reynolds, R. and Izurieta, C., 2009. Comparison of JSON and XML data interchange formats: a case study. Caine, 9, pp.157-162.

[83] O'Donnell, M., 1997. Variable-length on-line document generation. In the Proceedings of the 6th European Workshop on Natural Language Generation, Gerhard-Mercator University, Duisburg, Germany.

[84] Ono, K., Koyanagi, T., Abe, M. and Hori, M., 2002. XSLT stylesheet generation by example with WYSIWYG editing. In Proceedings 2002 Symposium on Applications and the Internet (SAINT 2002) (pp. 150-159). IEEE.

[85] Palmirani, M. and Governatori, G., 2018, December. Modelling Legal Knowledge for GDPR Compliance Checking. In JURIX (pp. 101-110).

[86] Palmirani, M. and Vitali, F., 2011. Akoma-Ntoso for legal documents. In Legislative XML for the semantic Web (pp. 75-100). Springer, Dordrecht.

[87] Parikh, A.P., Wang, X., Gehrmann, S., Faruqui, M., Dhingra, B., Yang, D. and Das, D., 2020. Totto: A controlled table-to-text generation dataset. arXiv preprint arXiv:2004.14373.

[88] Penadés, M.C., Martí, P., Canós, J.H. and Gómez, A., 2014. Product Line-based customization of e-Government documents. In PEGOV 2014: Personalization in e-Government Services, Data and Applications (Vol. 1181). CEUR-WS.

[89] Pérez, F. and Granger, B.E., 2007. IPython: a system for interactive scientific computing. Computing in Science & Engineering, 9(3), pp.21-29.

[90] Pérez, F. and Granger, B.E., 2017. The state of Jupyter: How Project Jupyter got here and where we are headed. Available at: https://www.oreilly.com/radar/the-state-of-jupyter (Accessed 15.5.2021).

[91] Perkel, J.M., 2018. Why Jupyter is data scientists' computational notebook of choice. Nature, 563(7732), pp.145-147.

[92] Pikus, Y., Weißenberg, N., Holtkamp, B. and Otto, B., 2019. Semi-automatic ontology-driven development documentation: generating documents from RDF data and DITA templates. In Proceedings of the 34th ACM/SIGAPP Symposium on Applied Computing (pp. 2293-2302).

[93] Pimentel, J.F., Murta, L., Braganholo, V. and Freire, J., 2021. Understanding and improving the quality and reproducibility of Jupyter notebooks. Empirical Software Engineering, 26(4), pp.1-55.

[94] Poore, G.M., 2019. Codebraid: Live Code in Pandoc Markdown.

[95] Priestley, M., Hargis, G. and Carpenter, S., 2001. DITA: An XML-based technical documentation authoring and publishing architecture. Technical communication, 48(3), pp.352-367.

[96] Radford, A., Narasimhan, K., Salimans, T. and Sutskever, I., 2018. Improving language understanding by generative pre-training.

[97] Radford, A., Wu, J., Child, R., Luan, D., Amodei, D. and Sutskever, I., 2019. Language models are unsupervised multitask learners. OpenAI blog, 1(8), p.9.

[98] Raffel, C., Shazeer, N., Roberts, A., Lee, K., Narang, S., Matena, M., Zhou, Y., Li, W. and Liu, P.J., 2019. Exploring the limits of transfer learning with a unified text-to-text transformer. arXiv preprint arXiv:1910.10683.

[99] Ragan-Kelley, B., Walters, W.A., McDonald, D., Riley, J., Granger, B.E., Gonzalez, A., Knight, R., Perez, F. and Caporaso, J.G., 2013. Collaborative cloud-enabled tools allow rapid, reproducible biological insights. The ISME journal, 7(3), pp.461-464.

[100] Reiter, E., 1995. NLG vs. Templates. arXiv preprint cmp-lg/9504013.

[101] Reiter, E. and Dale, R., 2000. Building Natural Language Generation Systems. Cambridge University Press.

[102] Reiter, E., Mellish, C. and Levine, J., 1995. Automatic generation of technical documentation. Applied Artificial Intelligence an International Journal, 9(3), pp.259-287.

[103] Reiter. E., 2016. NLG vs Templates: Levels of Sophistication in Generating Text. https://ehudreiter.com/2016/12/18/nlg-vs-templates/ (Accessed 15.5.2021).

[104] Rong, G., Jin, Z., Zhang, H., Zhang, Y., Ye, W. and Shao, D., 2019. DevDocOps: Towards automated documentation for DevOps. In 2019 IEEE/ACM 41st International Conference on Software Engineering: Software Engineering in Practice (ICSE-SEIP) (pp. 243-252). IEEE.

[105] Sewell, C., 2017, IPyPublish: A package for creating and editing publication ready scientific reports and presentations from Jupyter Notebooks. https://ipypublish.readthedocs.io/en/latest/ (Accessed 15.5.2021).

[106] Sifa, R., Ladi, A., Pielka, M., Ramamurthy, R., Hillebrand, L., Kirsch, B., Biesner, D., Stenzel, R., Bell, T., Lübbering, M. and Nütten, U., 2019. Towards automated auditing with machine learning. In Proceedings of the ACM Symposium on Document Engineering 2019 (pp. 1-4).

[107] Sripada, S., Burnett, N., Turner, R., Mastin, J. and Evans, D., 2014, June. A case study: NLG meeting weather industry demand for quality and quantity of textual weather forecasts. In Proceedings of the 8th International Natural Language Generation Conference (INLG) (pp. 1-5).

[108] Sutskever, I., Martens, J. and Hinton, G.E., 2011. Generating text with recurrent neural networks. In ICML.

[109] Sutskever, I., Vinyals, O. and Le, Q.V., 2014. Sequence to sequence learning with neural networks. arXiv preprint arXiv:1409.3215.

[110] Tang, J., Yang, Y., Carton, S., Zhang, M. and Mei, Q., 2016. Context-aware natural language generation with recurrent neural networks. arXiv preprint arXiv:1611.09900.

[111] Third, A., Williams, S. and Power, R., 2011. OWL to English: a tool for generating organised easily-navigated hypertexts from ontologies. In 10th International Semantic Web Conference (ISWC 2011), 23-27 Oct 2011, Bonn, Germany.

[112] Torre, D., Lachmann, A. and Ma'ayan, A., 2018. BioJupies: automated generation of interactive notebooks for RNA-Seq data analysis in the cloud. Cell Systems, 7(5), pp.556-561.

[113] Vaswani, A., Shazeer, N., Parmar, N., Uszkoreit, J., Jones, L., Gomez, A.N., Kaiser, L. and Polosukhin, I., 2017. Attention is all you need. arXiv preprint arXiv:1706.03762.

[114] Wen, E. and Weber, G., 2018. SwiftLaTeX: Exploring Web-based True WYSIWYG Editing for Digital Publishing. In Proceedings of the ACM Symposium on Document Engineering 2018 (pp. 1-10).

[115] Wiseman, S., Shieber, S.M. and Rush, A.M., 2017. Challenges in data-to-document generation. arXiv preprint arXiv:1707.08052.

[116] Yang, S., Wei, R. and Shigarov, A., 2018. Semantic interoperability for electronic business through a novel cross-context semantic document exchange approach. In Proceedings of the ACM Symposium on Document Engineering 2018 (pp. 1-10).

[117] Yang, S. and Wei, R., 2020. Semantic Interoperability Through a Novel Cross-Context Tabular Document Representation Approach for Smart Cities. IEEE Access, 8, pp.70676-70692.

[118] Yu, J., McCluskey, K. and Mukherjee, S., 2020. Tax Knowledge Graph for a Smarter and More Personalized TurboTax. arXiv preprint arXiv:2009.06103.

[119] Constantin, A., Peroni, S., Pettifer, S., Shotton, D. and Vitali, F., 2016. The document components ontology (DoCO). Semantic Web, 7(2), pp.167-181.

[120] Scheurer, J., Campos, J.A., Chan, J.S., Chen, A., Cho, K. and Perez, E., 2022. Training language models with natural language feedback. arXiv preprint arXiv:2204.14146.

[121] Choi, J.H., Hickman, K.E., Monahan, A. and Schwarcz, D., 2023. ChatGPT goes to law school. Available at SSRN.

[122] OpenAI, 2023. GPT-4 Technical Report. arXiv preprint arXiv:2303.08774

[123] Eloundou, T., Manning, S., Mishkin, P. and Rock, D., 2023. Gpts are gpts: An early look at the labor market impact potential of large language models. arXiv preprint arXiv:2303.10130.

[124] Schick, T., Dwivedi-Yu, J., Dessì, R., Raileanu, R., Lomeli, M., Zettlemoyer, L., Cancedda, N. and Scialom, T., 2023. Toolformer: Language models can teach themselves to use tools. arXiv preprint arXiv:2302.04761.

[125] Zhao, W.X., Zhou, K., Li, J., Tang, T., Wang, X., Hou, Y., Min, Y., Zhang, B., Zhang, J., Dong, Z. and Du, Y., 2023. A Survey of Large Language Models. arXiv preprint arXiv:2303.18223.

[126] Caufield, J.H., Hegde, H., Emonet, V., Harris, N.L., Joachimiak, M.P., Matentzoglu, N., Kim, H., Moxon, S.A., Reese, J.T., Haendel, M.A. and Robinson,

P.N., 2023. Structured prompt interrogation and recursive extraction of semantics (SPIRES): A method for populating knowledge bases using zero-shot learning. arXiv preprint arXiv:2304.02711.

[127] Chen, Z., Liu, Y., Chen, L., Zhu, S., Wu, M. and Yu, K., 2022. OPAL: Ontology-Aware Pretrained Language Model for End-to-End Task-Oriented Dialogue. arXiv preprint arXiv:2209.04595.

[128] Varshney, D., Zafar, A., Behera, N.K. and Ekbal, A., 2023. Knowledge grounded medical dialogue generation using augmented graphs. Scientific Reports, 13(1), p.3310.

[129] Li, F., UK, A., Hogg, D.C. and Cohn, A.G., 2022, November. Ontology Knowledge-enhanced In-Context Learning for Action-Effect Prediction. In Advances in Cognitive Systems. ACS-2022.